\documentclass{article}

\usepackage[english]{babel}
\usepackage{comment}
\usepackage{longtable, booktabs}
\usepackage[letterpaper,top=2cm,bottom=2cm,left=3cm,right=3cm,marginparwidth=1.75cm]{geometry}

\usepackage{amsmath}
\usepackage{graphicx}
\usepackage{float}
\usepackage[colorlinks=true, allcolors=blue]{hyperref}
\usepackage{url}
\usepackage[official]{eurosym}
\usepackage{authblk}

\title{A real-time Artificial Intelligence system for learning Sign Language}

\author[1]{Elisa Cabana \thanks{Corresponding author's email: \href{mailto:elisa.cabana@cunef.edu}{elisa.cabana@cunef.edu}}}
\affil[1]{CUNEF University, Madrid, Spain}

\date{}

\begin{document}
\maketitle

\begin{abstract}
A primary challenge for the deaf and hearing-impaired community stems from the communication gap with the hearing society, which can greatly impact their daily lives and result in social exclusion. To foster inclusivity in society, our endeavor focuses on developing a cost-effective, resource-efficient, and open technology based on Artificial Intelligence, designed to assist people in learning and using Sign Language for communication. The analysis presented in this research paper intends to enrich the recent academic scientific literature on Sign Language solutions based on Artificial Intelligence, with a particular focus on American Sign Language (ASL). This research has yielded promising preliminary results and serves as a basis for further development.
\end{abstract}

\section{Introduction}

Over $5\%$ of the global population, i.e., approximately more than 430 million people, experiences hearing impairments according to the World Health Organization (WHO) \cite{who}. Of them, 34 million are children. These numbers are anticipated to rise due to aging demographics and increased exposure to noise. WHO estimated that by 2050 over 700 million people - or 1 in every 10 people - will have disabling hearing loss. A substantial portion faces significant impairments from childhood, and sign languages serve as an effective tool to overcome this communication barrier. Collectively, they use more than 300 different sign languages. Despite the prevalence of sign languages within the deaf community, challenges persist in interactions with hearing individuals, highlighting the need for improved communication solutions.

In the United States (US), approximately 2 to 3 out of every 1,000 children are born with a detectable hearing loss in one or both ears, according to the National Institute on Deafness and Other Communication Disorders (NIDCD) \cite{NIDCD}. While it is difficult to determine the exact number, it is estimated that approximately more than a half-million people throughout the US use ASL to communicate as their native language \cite{mitchell2006many}. ASL is the third most commonly used language in the United States, after English and Spanish according to the Commission on the Deaf and Hard of Hearing \cite{CDHH}.

A communication gap separates the hearing population from the deaf community, frequently hindering their ability to engage fully in an equal society, which can result in social, cultural, and even workplace marginalization. This research paper aims to tackle these issues, by creating accessible open technologies based on Artificial Intelligence that are cost-effective, affordable and require minimal resources for their use. Specifically, the study aims to create a computer vision system for American Sign Language (ASL) recognition and translation in real-time that can serve as a learning application. For this purpose, an extensive dataset of images of ASL alphabet signs has been compiled. We studied several neural networks classification models and implemented into the final system the one with the overall best performance metrics. This provides valuable insights for the sign language translation scenario. Our system can act as an affordable solution for the learning process of both deaf and hearing individuals in situations where interpreters are unavailable, specially for children with hearing disability who want or need to reinforce what they learn at the school. This contribution significantly advances ongoing academic research on Artificial Intelligence-based Sign Language solutions, specifically focusing on American Sign Language (ASL).

\section{Motivation and background}

Over the past decade, Sign Language recognition and translation has emerged as a prominent area of research. Numerous approaches have been explored in this field. The problem is that most of them are based on costly hardware or resources, such as the use of gloves with specific sensors \cite{waldron1995isolated, kadous1996machine, vogler1997adapting, lokhande2015data, kute2020} to track hand movements, systems based on a Kinect sensor to extract the 3D skeleton \cite{kumar2018position}, or a combination of cameras \cite{starner1997real}. In the market there are also other solutions, such as, StorySign which is a mobile application developed by Huawei that uses augmented reality technology to translate children's books into sign language \cite{sto}. Content4All is another initiative, funded by the European Union under the Horizon 2020 framework, aimed at improving accessibility to media content for people with disabilities, particularly those with sensory impairments such as visual or hearing impairments \cite{c4a}. The project focuses on developing technologies for automatically generating sign language avatars and audio descriptions to make television, film, and other media content more accessible to a wider audience. SignAll is a company that has developed software and hardware solutions that utilize Artificial Intelligence and Computer Vision technologies to interpret Sign Language gestures and translate them into spoken language or text \cite{sal}. Their pricing structure may vary depending on the specific solution and the level of customization or support required, for example SignAll Chat which is their real-time sign language translation tool, have subscription fees associated with it, especially for businesses or organizations that integrate it into their communication systems. Similarly, SignAll Learn, their educational platform for learning Sign Language, might offer certain free resources or basic lessons but may require payment for access to premium content or features. These solutions demonstrate significant effort in this area, but they typically come with a hefty price tag, as they often involve advanced technologies, ongoing development, and support, making these systems impractical for cost-free real-world applications.

On the other side, Sign Language is challenging due to the diversity of languages across different countries, such as Lengua de Signos Española (LSE) in Spain, Lingua dei Segni Italiana (LIS) in Italy, and American Sign Language (ASL) in the USA, among others. Creating a universal Sign Language recognition model is a complex task. Nonetheless, there are some cost-effective proposed solutions in the literature for certain languages \cite{pigou2015sign, kamruzzaman2020arabic, rastgoo2021sign, triwijoyo2023deep, rodriguez2022hierarchical, morillas2023sign4all}, but most of them have some limitations, e.g. lacking of high accuracy for identifying several classes. This gap highlights the need for strengthening development efforts in this particular domain.

\section{Data}

For training a neural network model for sign language recognition, one would need a dataset consisting of images of hands displaying various signs. Given that certain words are directly derived from the manual alphabet \cite{baez2016colours}, we collected data on the American Sign Language (ASL) alphabet. 

The task of finding a well-curated dataset for tasks like sign language recognition and translation is a challenging and time-consuming process, with no guarantee of success. Because ASL is a widely used language, we were able to discover some free datasets online, but they lacked quality. Therefore, we decided to invest time in creating a diverse set of high-quality images for the task. Nonetheless, generating these images requires significant effort and effective research. Some challenges are that the dataset should contain a diverse set of images capturing different hand gestures, angles, positions, proximity, lighting conditions, and backgrounds. The reason is that variability in the images helps the model generalize better to unseen data, and this is extremely important specially for the usability in real-time where the scenario around the user can be highly variable. Another challenge is the annotation, since each image in the dataset should be labeled with the corresponding sign it represents. 

Our data collection process consists of three stages. In the first stage, a set of videos for each sign are recorded at 30 frames per second, capturing a continuous stream of information. Different locations, backgrounds and hand positions are considered. In the second stage, the videos are processed through a Python script that extracts an image from each video every 2 frames. The script saves the images in separate folders for each letter, and assigns as the folder name the corresponding letter. In the last stage all images are subjected to a manual post-processing step to remove invalid ones, to ensure that the final set of images were of high quality with clear visibility of hand gestures, since low-resolution or blurry images may hinder the model's ability to learn meaningful features. This reduced the original sample size for each class, but increased the quality of the training dataset.

Nonetheless, in the future, we expect to evaluate other alternatives for data collection such as increasing the time of extraction, as well as increasing locations, with the aim of reducing redundancy, adding more variability to the data and optimizing the use of computational resources. 

\begin{figure}[H]
    \centering
    \includegraphics[width=0.7\linewidth]{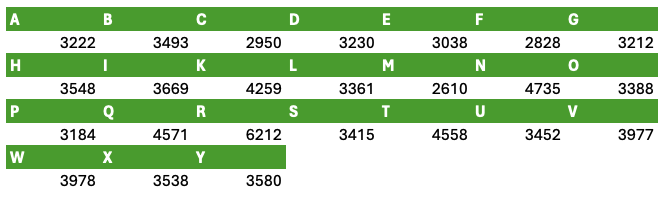}
    \caption{\label{fig:samplesize}Sample Size for the ASL alphabet dataset.}
\end{figure}

Using this method we were able to collect a dataset covering the various static ASL letters, ensuring that there is a sufficient number of samples for each class to avoid class imbalance issues. The final training dataset consists of 88K images of size $192 \times 192$. The alphabet encompasses both static and dynamic signs, often used for proper nouns, streets, trademarks, or words lacking a designated sign or whose sign is unknown. In this paper, the dynamic letters J and Z which require hand movements were excluded and left for future work. Figure \ref{fig:samplesize} shows the sample size for each letter.

During the model training process we also considered ``Data Augmentation'', by applying transformations such as small rotations, scaling, translation, and flipping to increase the variability of the training data, and ultimately improve the model's robustness reducing overfitting.

A validation dataset of 720 images was created for tuning the hyper-parameters and monitor performance during training. It consists of different images collected in different scenarios to evaluate the model's generalization ability on unseen data. From the beginning, the data collection process was an iterative approach that involved continuous refinements and improvements through multiple rounds in order to address the deficiencies and enhancing the dataset quality.


\section{Model development and evaluation performance}

\subsection{Model definition}

\begin{figure}[H]
    \centering
    \includegraphics[width=0.85\linewidth]{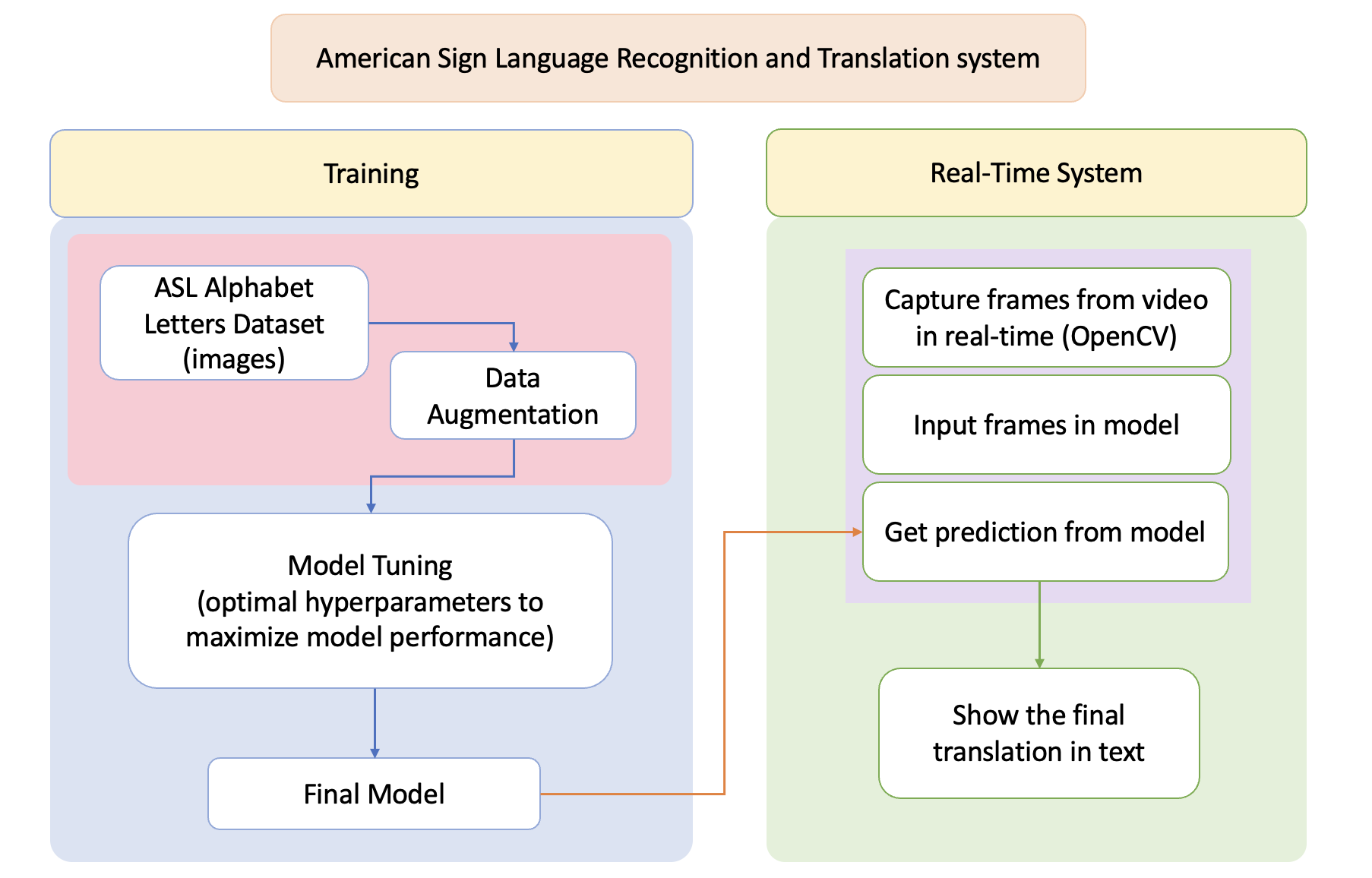}
    \caption{\label{fig:schema}Pipeline of the ASL recognition and translation system.}
\end{figure}

Once the dataset was ready, we were able to delve into the model training process. Figure \ref{fig:schema} shows the pipeline of the proposed ASL recognition and translation system. It consists of two stages: the first one for training different models and selecting the most optimal one, balancing accuracy and computational efficiency. In the second stage, a real-time computer vision system was developed, capable of identifying what is signed in the screen in real-time and translating this into English text. The system captures the frames from the camera of the device using the OpenCV library, then inputs each frame into the final model, gets the sign classification and shows it in the user screen translated into English text.



From a research perspective, there are several deep learning architectures that can be used for the Sign Language classification task. One particular neural network technique, specially useful for this purpose is the Convolutional Neural Network (CNN) model, which have shown significant success in image recognition. This method is inspired by the human visual cortex, since it employs multiple layers performing discrete convolutions, activation functions, and operations like pooling to achieve classification. We explored 8 different types of pre-defined Convolutional Neural Network families, each with different model configurations. The CNN families discussed in this paper are as follows:

\subsubsection{VGG}

VGG, short for Visual Geometry Group, is a standard deep Convolutional Neural Network architecture with multiple layers. VGG was proposed by A. Zisserman and K. Simonyan from the University of Oxford in 2014 \cite{simonyan2014very}. This model is known for its simplicity and uniformity in architecture. It consists of a series of convolutional layers, each followed by a max-pooling layer, with fully connected layers at the end. The convolutional layers use small 3x3 filters with a stride of 1 and same padding. This design choice enables VGG to incorporate a large number of layers, enhancing its performance. The model has been used in two versions, one with 16 layers (13 convolutional layers and 3 fully connected layers), and the other one with 19 (16 convolutional layers and 3 fully connected layers). VGG architectures are widely used and have served as a baseline for deeper convolutional neural network architectures. Their simplicity and effectiveness make them a popular choice for various computer vision tasks.

\subsubsection{Inception}

The Inception network family, developed by Google researchers in 2015, encompasses several versions, each marked by significant advancements \cite{szegedy2015going}. The name comes from the popular movie ``Inception'', which revolves around the idea of multiple layers of reality. Similarly, Inception networks consist of multiple layers, each with its own set of convolutions. The initial version, Inception V1, unveiled in 2015 as GoogLeNet, boasted 7 million parameters. This version introduced the groundbreaking ``inception module'', which are multi-branch convolutional modules designed to capture features at different spatial scales. Each inception module typically consists of convolutions of different filter sizes (1x1, 3x3, 5x5) and pooling operations, allowing the network to capture both local and global features effectively. Inception modules often include 1x1 convolutions for dimensionality reduction. These 1x1 convolutions help reduce the number of input channels, thereby reducing computational complexity while preserving information. Inception V2, a subsequent version, incorporated batch normalization and other architectural enhancements. Inception V3 further refined the model by incorporating factorization ideas in convolution layers, effectively reducing dimensionality and mitigating overfitting. This innovation led to a remarkable reduction in parameters. Additionally, Inception V3 introduced an efficient grid size reduction technique, enhancing computational efficiency without sacrificing performance. The final iteration, Inception V4 (also known as Inception-ResNet), integrated residual connections akin to ResNet's design. These connections enhanced the flow of information within the network.

\subsubsection{ResNet}

Residual Network (ResNet) stands out as one of the most widely acclaimed deep neural networks for image classification \cite{szegedy2016rethinking, he2016deep}. Unlike its predecessors that enhanced performance by increasing network depth, ResNet introduced the innovative concept of identity connections between layers, leading to the creation of residual blocks. These connections allow information to bypass one or more layers, generating identity maps. Notably, these connections do not introduce additional parameters or computational complexity. Consequently, they prevent an escalation in model training errors as the network deepens.

A residual block introduces a shortcut connection (skip connection) that bypasses one or more layers. This skip connection allows the model to learn residual functions, making it easier to train very deep networks. The skip connection in a residual block aims to learn the identity mapping, i.e., the output of the block is the sum of the input and the output of the block's internal layers. This helps in addressing the vanishing gradient problem. ResNet comes in various architectural variants, denoted as ResNet-18 up to ResNet-200. The number in the variant name represents the total number of layers in the network. The deeper variants have more residual blocks. Deeper ResNet variants use a ``bottleneck" architecture in which each residual block consists of three convolutional layers: 1x1, 3x3, and 1x1. This design reduces computational complexity while retaining representational power. ResNet typically uses global average pooling (GAP) instead of fully connected layers at the end of the network. GAP helps reduce the number of parameters and encourages the model to focus on the most important features.


ResNet has been widely used for various computer vision tasks, including image classification, object detection, and image segmentation. Pre-trained ResNet models on large image datasets (e.g., ImageNet) are often used as a starting point for transfer learning in other computer vision tasks. The learned features in these models can be fine-tuned for specific tasks with smaller datasets. ResNet has had a significant impact on the field of deep learning, and its principles of residual learning and skip connections have influenced the design of subsequent neural network architectures.


\subsubsection{Xception}

Xception is a neural network introduced by François Chollet, the creator of Keras, in 2017 \cite{chollet2017xception}. Its name stands for ``Extreme Inception," indicating its inspiration from the Inception architecture. However, it deviates from Inception's approach by assuming that cross-channel correlations and spatial correlations in feature maps can be entirely decoupled. In Xception, the key innovation is the depthwise separable convolution. Unlike traditional convolutions that operate on the entire input volume, depthwise separable convolutions break down the convolution operation into two steps: depthwise convolutions and pointwise convolutions. This separation reduces the computational complexity since it is more parameter-efficient than traditional convolutions, while maintaining representational capacity. Xception's overall architecture follows a similar structure to the Inception family of models. It employs multiple convolutional blocks, each containing depthwise separable convolutions, batch normalization, and rectified linear unit (ReLU) activations. In essence, Xception can be described as a linear stack of depthwise separable convolution layers, incorporating residual connections similar to those found in ResNet architectures. Xception has been primarily used for image classification tasks, but its principles can be applied to various computer vision tasks, including object detection and image segmentation. It is particularly effective in scenarios where computational efficiency is crucial.

\subsubsection{DenseNet}

DenseNet, short for ``Dense Convolutional Network'', is a deep neural network architecture introduced in 2017 \cite{huang2017densely}. It is designed to address some of the limitations of traditional convolutional neural networks by fostering stronger feature reuse and alleviating the vanishing-gradient problem. The core idea behind DenseNet is the concept of dense connectivity, where each layer is connected to every other layer in a feed-forward fashion. Unlike traditional CNNs where each layer is connected only to its subsequent layer, DenseNet creates dense connections between layers within a dense block, in which each layer receives feature maps from all preceding layers as input and passes its feature maps to all subsequent layers. To reduce computational complexity and model size, DenseNet typically employs bottleneck layers within dense blocks. This design choice helps in improving computational efficiency while preserving representational capacity. Between dense blocks, DenseNet includes transition layers that perform down-sampling operations. Transition layers reduce the spatial dimensions of feature maps and control its growth through the network.

\subsubsection{MobileNet}

MobileNet is a family of lightweight convolutional neural network architectures specifically designed for efficient inference on mobile and embedded devices with limited computational resources. It was introduced in 2017 \cite{howard2017mobilenets}. MobileNet primarily employs depthwise separable convolutions, which significantly reduces the computational cost and model size while preserving representation capacity. Its architectures offer flexibility through two hyperparameters: width multiplier and resolution multiplier. The width multiplier controls the number of channels in each layer, allowing for trade-offs between model size and accuracy. The resolution multiplier scales down the input resolution, further reducing computational complexity. This architecture is more suitable for real-time applications on resource-constrained devices, making them popular choices for mobile vision applications.

MobileNetV2 is an improved version of the original MobileNet architecture, designed to further enhance efficiency and performance for mobile and embedded vision applications. It was introduced by Mark Sandler et al. from Google in 2018 \cite{sandler2018mobilenetv2}. MobileNetV2 introduces the concept of inverted residuals, which consist of lightweight depthwise separable convolutions followed by linear bottlenecks. This reduces computational complexity while preserving representational capacity, leading to improved performance. Linear bottlenecks are added between layers to increase the non-linearity of the network, allowing for better feature learning and representation. These bottlenecks help alleviate the issue of information bottlenecking, which can occur in shallow networks with depthwise separable convolutions. MobileNetV2 employs expansion and projection layers within each inverted residual block. The expansion layer increases the number of channels, while the projection layer reduces the dimensionality back to the desired number of channels. This design enhances the expressive power of the network without significantly increasing computational cost.

MobileNetV3 is the third iteration of the MobileNet architecture, and it was introduced in 2019 \cite{howard2019searching}. It represents a significant advancement over previous versions, leveraging Neural Architecture Search (NAS) techniques to design more efficient and accurate convolutional neural network (CNN) architectures optimized for mobile and embedded devices.

\subsubsection{NASNet}
NASNet, which means ``Neural Architecture Search Network'', is a family of neural network architectures introduced in 2018 \cite{zoph2018learning}. Neural Architecture Search (NAS) techniques involves searching through a large space of possible architectures to find configurations that optimize specific objectives, such as accuracy, efficiency, or computational cost. That is why NASNet is considered as being automatically designed through reinforcement learning or evolutionary algorithms rather than hand-crafted by human experts. The search space explored by NASNet includes various architectural decisions such as the number of layers, types of operations (e.g., convolution, pooling), skip connections, and other structural components. It is designed to be efficient and scalable, capable of automatically discovering architectures that achieve state-of-the-art performance on image recognition tasks while being computationally efficient.

NASNet could be incorrectly confused with MobileNetV3 since both are designed using NAS techniques, which automate the process of discovering optimal network architectures through exploration of a large design space. But NASNet architectures are typically the result of extensive experimentation and optimization, while MobileNetV3 leverages hand-designed architectural innovations along with insights from NAS, i.e., it incorporates some elements inspired by NAS, but its design is not solely driven by it.

\subsubsection{RegNet}

RegNet is another family of neural networks and it was introduced in 2020 \cite{radosavovic2020designing}. RegNet architectures are designed with a focus on providing highly regularized structures that offer a wide range of computational efficiency and accuracy trade-offs. These regularization principles aimed at controlling overfitting and improving generalization performance. The design space of RegNet encompasses a broad range of network configurations, including variations in depth, width, and resolution. By exploring this design space, the models can be tailored to specific computational budgets and performance requirements. RegNet architectures are designed to be scalable and transferable across different tasks and domains. They can be fine-tuned or adapted for various computer vision applications. Within the RegNet family, variants such as RegNetX and RegNetY have been introduced to explore different aspects of the design space and further refine the regularization principles. These variants offer additional flexibility and versatility, which allows to explore different aspects of the design space, fine-tune regularization techniques, and tailor neural network architectures to specific application requirements and constraints.

\subsection{Model results}

We trained the previously described families of convolutional neural networks models. We use the online tool NeptuneAI for keeping a track record and compare all models according to accuracy and loss in train and validation datasets, computational efficiency, and specific hyper-parameters. 

Table \ref{table:models} shows a summary about the top 10 models with the best performance. In general, all models were trained on a GPU NVIDIA Tesla T4 and executed on a GPU RTX3090. All of them also have the following common parameters: maximum number of epochs 50, batch size of 64, learning rate of 0,001 and SGD optimizer.

\begin{table}[H]
  \caption{Top 10 models' performance comparison.}
        \begin{tabular}{llllllll}
        
\midrule

Model  & Epochs &  Train   &  Train  &  Validation &  Validation & Time  &  Time   \\
   
Name &  Executed &   Accuracy &  Loss  &  Accuracy & Loss  & Train &  Execution   \\
\midrule
DenseNet201 & 47	& 0,9998	& 0,0483	& 	0,8042		&  1,2051		&  3h 44m		&  37ms \\
DenseNet169	 & 39	 & 0,9998	 & 0,0876	 & 0,7931	 & 1,2188	 & 2h 29m	 & 29ms \\
RegNetY064		 & 50		 & 0,9996		 & 0,0544		 & 0,7917		 & 1,1983		 & 4h 40m		 & 36ms \\
ResNet152		 & 50		 & 0,9998		 & 0,0367		 & 0,7833	 & 	1,2994		 & 5h 19m 	 & 	48ms \\
RegNetX040		 & 48		 & 0,9995		 & 0,0631		 & 0,7764		 & 1,1877		 & 3h 02m		 & 25ms \\ 
InceptionV3	 & 	43		 & 0,9995		 & 0,0708	 & 	0,7667		 & 1,3883		 & 1h 24m		 & 14ms \\
MobileNetV2 	 & 	39		 & 0,9995		 & 0,0905		 & 0,7542		 & 1,3007		 & 59m &	10ms \\
NASNet		 & 49		 & 0,9996		 & 0,0421	 & 	0,7542		 & 1,7506		 & 10h 12m		 & 79ms \\
Xception		 & 50		 & 0,9992		 & 0,0518		 & 0,7153		 & 1,4442		 & 2h 51m		 & 22ms \\
VGG16		 & 30		 & 0,9993		 & 0,1762		 & 0,6889		 & 2,2666		 & 1h 42m		 & 25ms \\ 
	\end{tabular}%
  \label{table:models}%
\end{table}%


Note that the rows in the table are sorted in decreasing order by the validation accuracy values. The number of epochs executed sometimes is lower than the fixed maximum of 50 because we use Early Stopping, a technique used to prevent overfitting during the training of a model. It involves monitoring a specified metric (such as validation loss or accuracy) during the training process and stopping the training when the metric stops improving or starts deteriorating, even before the model has completed all epochs. 

The top 10 models have a validation accuracy ranging from $68\%$ to $80\%$. Although, NASNet is the most computationally inefficient when training (10h) and executing (79ms). DenseNet201 is the model with the highest validation accuracy, while maintaining a low validation loss, as well as good performance metric in the training set. Although, there are more efficient alternatives in terms of training and execution time, such as InceptionV3 and MobileNetV2, which has around $5-7\%$ less validation accuracy but with the advantage of having around $70\%$ training and execution time improvement. Therefore, depending on the final use-case, one model could be preferable versus another, and it is recommended that for a real-time translation task, more computationally efficient models such as Inception or MobileNet could be used.

\section{Conclusions and Future Work}

For our real-time system we decided to use the DenseNet201 model, since at the moment it is designed to be a learning application that does not require such demanding computational efficiency. The online tool can be tested from a computer or a mobile phone in the following link: \url{https://signlanguagerecognition.aprendeconeli.com}. However, while the accuracy level achieved by the DenseNet201 model is commendable given the complexity of the task, there is still room for improvement, especially with enhanced computational resources dedicated to training the models.

For future work, our immediate priority is improving the computational resources to expand the comparative study and test other models and approaches that right now are truly difficult to explore. Adequate resources will help us accelerate the training process and it will also empower us to investigate complex model architectures, such as heavy models that we could not explore before, or techniques like ``Transfer Learning''. The latter aims to leverage the knowledge gained from solving one problem (such as image classification using ImageNet) and applies it to a different but related problem (such as our Sign Language recognition and classification task). In other words, instead of training a model from scratch, the pre-trained model's learned features are reused as a starting point for the new task. The early layers of the pre-trained model, which capture low-level features like edges and textures, are typically retained, while the later layers may be replaced or fine-tuned, adapting the model to the new task by using task-specific data, allowing it to learn the  patterns and nuances that are actually relevant for the specific problem. Transferring knowledge can help improve the performance of the model and therefore it is a very interesting path we aim to explore.

Additionally, we would also like to investigate other paths in order to improve the system and make it practical in the future. On the technical front, we aim to explore background segmentation techniques. The optimization in this area is pivotal for ensuring the system's robustness across varied environmental settings, guaranteeing seamless performance and reliability in diverse contexts where the technology may be utilized. To further refine the AI system's capabilities, we aim to explore alternative paths such as the hand key-points extraction. This advancement might significantly contribute to improving the system's precision in recognizing intricate gestures, thereby enhancing its overall accuracy and effectiveness. Moreover, a critical forthcoming initiative involves the development of an engaging graphic interface, potentially tailored for interactive Sign Language learning. The convergence of these future endeavors is poised to propel our technology towards greater inclusivity, efficiency, and effectiveness in breaking down communication barriers within the Sign Language community. Through persistent innovation and collaborative efforts, we aim to contribute with our grain of sand to forge a more inclusive societal landscape where communication barriers are minimized, fostering a more connected and supportive environment for all.

\bibliographystyle{unsrt}
\bibliography{sample}

\end{document}